\documentclass{article} 
\usepackage{iclr2026_conference,times}

\usepackage{amsmath,amsfonts,bm}

\def\eqref#1{equation~\ref{#1}}

\def\1{\bm{1}}

\DeclareMathAlphabet{\mathsfit}{\encodingdefault}{\sfdefault}{m}{sl}
\SetMathAlphabet{\mathsfit}{bold}{\encodingdefault}{\sfdefault}{bx}{n}

\usepackage{hyperref}
\usepackage{url}

\usepackage{algorithm}
\usepackage{algorithmic}
\usepackage{graphicx}
\usepackage{xcolor} 
\usepackage{multirow}
\usepackage{array}
\usepackage{wrapfig}
\usepackage{xcolor}
\usepackage{subcaption} 
\usepackage{amsfonts}
\usepackage{booktabs}

\usepackage{wrapfig}
\usepackage{array}

\title{Entropy-guided $k$-Guard Sampling for Long-Horizon Autoregressive Video Generation}

\author{%
\parbox{\textwidth}{\centering\normalfont\mdseries
\vspace{18pt}
Yizhao Han$^{1,\,2*}$ \quad
Tianxing Shi$^{1*}$ \quad
Zhao Wang$^{3}$ \quad
Zifan Xu$^{1}$ \quad
Zhiyuan Pu$^{3}$\\[2pt]
Mingxiao Li$^{2}$ \quad
Qian Zhang$^{2}$ \quad
Wei Yin$^{2\ddagger}$ \quad
Xiao-Xiao Long$^{1\dagger}$\\[4pt]
$^{1}$Nanjing University \qquad
$^{2}$Horizon Robotics \qquad
$^{3}$China Mobile\\[4pt]
\url{https://greanguy.github.io/ENkG}
}%
}

\iclrfinalcopy

\begin{document}
\maketitle

\begin{abstract}

Autoregressive (AR) architectures have achieved significant successes in LLM, inspiring explorations for video generation.  In LLMs, top-$p$/top-$k$ sampling strategies work exceptionally well: language tokens have high semantic density and low redundancy, so a fixed size of token candidates already strike a balance between semantic accuracy and generation diversity. In contrast, video tokens have low semantic density and high spatio-temporal redundancy. This mismatch makes static top-k/top-p strategies ineffective for video decoders: they either introduce unnecessary randomness for low-uncertainty regions (static backgrounds) or stuck in early errors for high-uncertainty regions (foreground objects). Prediction errors will accumulate as more frames are generated and eventually
severely degrade long-horizon quality. 
To address this, we propose Entropy-Guided $k$-Guard (ENkG) sampling, a simple yet effective strategy that adapts sampling to token-wise dispersion, quantified by the entropy of each token’s predicted distribution. 
ENkG uses adaptive token candidate sizes: for low-entropy regions, it employs fewer candidates to suppress redundant noise and preserve structural integrity; for high-entropy regions, it uses more candidates to mitigate error compounding.
ENkG is model-agnostic, training-free, and adds negligible overhead. Experiments demonstrate consistent improvements in perceptual quality and structural stability compared to static top-k/top-p strategies.

\end{abstract}

\section{Introduction} 
\label{sec:intro}
The field of video-based world models has witnessed explosive growth in recent years, with significant advancements in generating high-fidelity, temporally coherent, and physically plausible video sequences \cite{villegas2022phenaki, wang2023worlddreamer}. These models aim to build an internal representation of the world's dynamics, enabling applications from realistic simulation for robotics to advanced content creation \cite{he2025pretrained}. This progress has paved the way for models that can not only synthesize video from text but also begin to understand and simulate interactive environments \cite{mo2025dreamland}.

\label{sec:motivation}
\begin{figure}[tp]
  \centering
  \includegraphics[width=\linewidth]{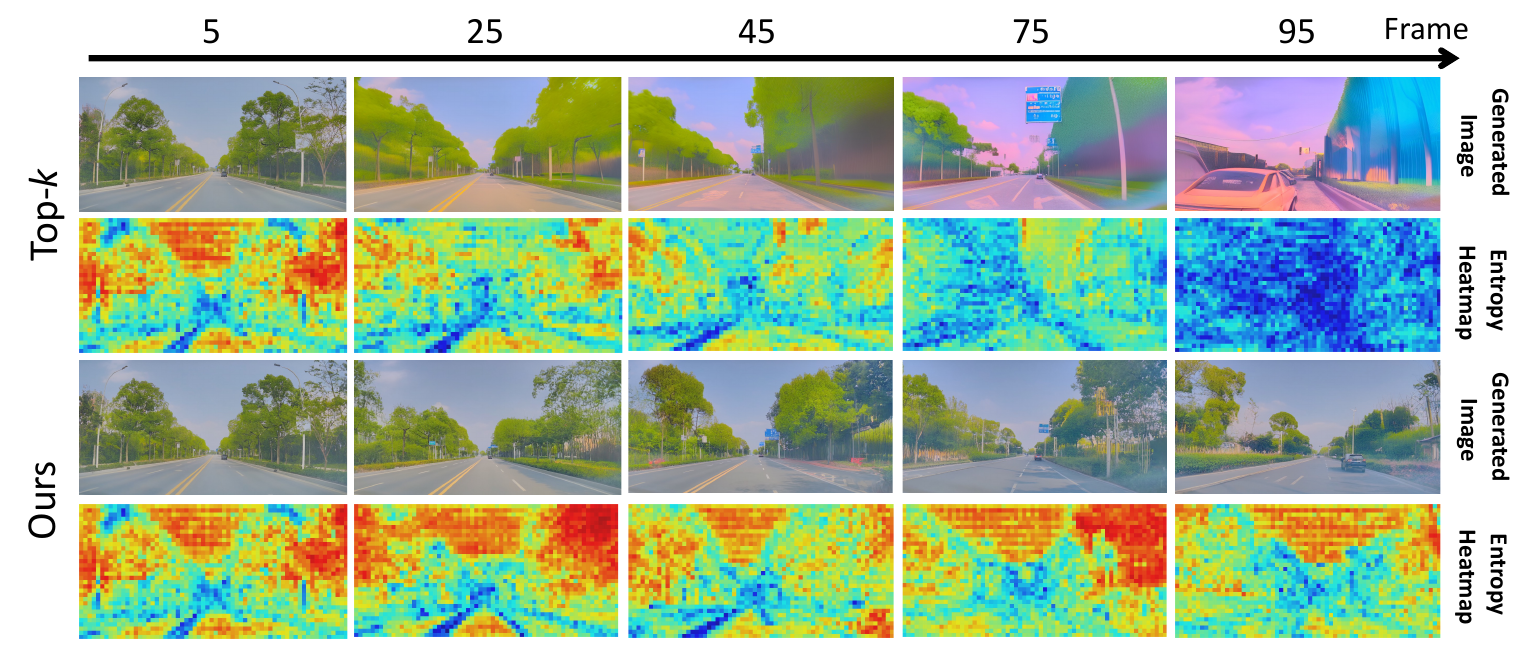}
  \caption{
  The results illustrate the phenomenon of entropy collapse in standard AR decoding, where blue regions indicate low entropy and red regions indicate high entropy. Our method effectively alleviates this issue.}
  \label{fig:entropy_spread}
  \vspace{-6mm}
\end{figure}

Among the various architectural paradigms, autoregressive (AR) models have become a cornerstone for video generation. By factorizing the joint probability distribution of video frames into a product of conditional probabilities, AR models excel at capturing temporal causality and allow for fine-grained, frame-by-frame control during generation. This sequential approach is inherently flexible, supporting variable-length video generation and compatibility with scalable transformer architectures~\citep{dosovitskiy2020image, weissenbornscaling}. However, the sequential nature of AR models also introduces significant challenges: error accumulation~\citep{parthipan2024defining,bengio2015scheduled} and exposure bias~\citep{schmidt2019generalization}. Minor inaccuracies or suboptimal choices in generating a single frame can propagate and amplify over time, leading to a degradation of quality, loss of coherence, and "drifting" from the intended content in longer video sequences~\citep{hu2023gaia}.



Several strategies have been proposed to mitigate this effect. \citet{huang2025selfforcingbridgingtraintest} simulate inference during training by feeding the model its own previous predictions, allowing it to learn to correct mistakes. Other works introduce noisy or masked contexts, encouraging the model to be robust to imperfect inputs~\citep{ren2025xar, zhou2025magi}. While effective, these approaches often require modifications to the model architecture or additional training complexity, which may limit their applicability to existing large-scale video generation models.

In contrast, we focus on the often-overlooked role of the \textbf{\emph{sampling process}} in autoregressive video generation. Our analysis reveals that conventional strategies such as fixed top-$k$ or nucleus (top-$p$) sampling fail to account for the spatially structured uncertainty inherent in video tokens. Specifically, we observe that high-entropy regions, corresponding to complex textures like foliage or road markings, are prone to brittleness, whereas low-entropy regions representing structured geometry can suffer from overconfidence and texture wash-out. This motivates an \textbf{\emph{adaptive}} sampling policy that modulates token diversity based on entropy, effectively balancing stability and richness in generated content.

Specifically, we introduce \textbf{Entropy-guided $k$-Guard sampling}, a model-agnostic algorithm that dynamically adjusts the size of candidates for each token according to its entropy. 
First, our method measure the entropy of each video token that indicates the dispersion of video token probability.
For low-entropy regions, our strategy employs fewer candidates to suppress redundant noise and preserve structural integrity; for high-entropy regions, it uses more candidates to mitigate error compounding.
Unlike previous solutions that modify training or rely on multiple candidate evaluations, our approach operates purely at the inference stage, making it widely applicable to existing autoregressive video models.  

We validate our method on several state-of-the-art autoregressive video generation architectures, demonstrating that it significantly reduces error accumulation, preserves fine-grained textures, and improves temporal coherence over extended sequences. Quantitative metrics and qualitative results confirm that our adaptive sampling strategy enables longer, more realistic video generation without retraining or architectural changes. These findings suggest that carefully designed inference-time strategies can be a powerful tool for improving autoregressive video generation, complementing existing advances in model design and training.

In summary, our contributions are threefold: (i) we identify the limitations of fixed sampling strategies in autoregressive video generation and highlight the role of spatially structured uncertainty in error accumulation, (ii) we propose a simple yet effective entropy-guided adaptive sampling strategy with a $k$-guard mechanism, and (iii) we empirically demonstrate that this method improves long-sequence video quality across multiple benchmark models. This work highlights the potential of uncertainty-aware inference as a practical and generalizable solution for high-fidelity video synthesis.

\section{Related Work}
\subsection{Video World Models} 
Video-based world models aim to learn an internal representation of an environment, allowing the system to predict future states, simulate interactions, and support planning~\citep{ha2018world, ding2024understanding, long2025surveylearningembodiedintelligence, zhang2025epona}. Recent progress in large-scale video generation has enabled the creation of high-fidelity world simulators capable of producing visually realistic and physically plausible sequences~\citep{openai2024video}, which are particularly valuable for applications such as autonomous driving and robotics~\citep{li2025roboticworldmodelneural}.

A wide range of generative architectures have been explored for video synthesis. Diffusion models and Generative Adversarial Networks (GANs) have shown success in producing high-quality frame sequences~\citep{ho2022video}, while autoregressive (AR) models have emerged as a powerful alternative due to their inherent capacity to model temporal dependencies. By factorizing the joint distribution of video tokens or frames into a product of conditional probabilities, AR models generate coherent, long-duration videos either frame-by-frame~\citep{gu2025long} or token-by-token~\citep{wu2024ivideogpt}, enabling fine-grained temporal control and interactive generation.

Despite their strengths, AR models suffer from a fundamental limitation: \emph{error accumulation}~\citep{bengio2015scheduled, ross2011reductionimitationlearningstructured}. During inference, each predicted frame or token is fed back as input for subsequent steps, so any imperfection can propagate and amplify over time~\citep{bengio2015scheduled, huang2025selfforcingbridgingtraintest}. This leads to quality degradation, manifesting as flickering, unnatural motion, or drift from the intended scene~\citep{kong20253d}. Error accumulation has been identified as a core challenge in theoretical analyses of autoregressive video generation, alongside issues such as memory bottlenecks~\citep{wang2025erroranalysesautoregressivevideo}.

To address this limitation, our work proposes a \emph{token-level adaptive sampling} strategy that dynamically modulates the sampling distribution according to the model's predictive uncertainty. By integrating uncertainty into the generation process, we directly mitigate the compounding of errors, preserving both temporal coherence and visual fidelity in long-duration video sequences.

\subsection{Autoregressive Sampling Algorithms} 
Sampling strategies are central to autoregressive (AR) generation. 
\emph{Greedy decoding} selects the most likely token at each step, but often yields low diversity. 
\emph{Beam search} explores multiple hypotheses in parallel and improves likelihood-based metrics, yet typically reduces diversity. 
A widely used family of stochastic methods is \emph{truncated probability sampling}, including top-$k$ ~\citep{noarov2025foundationstopkdecodinglanguage} and nucleus (top-$p$) sampling \cite{ravfogel2023conformal}. Both restrict sampling to a subset of the distribution, balancing diversity and quality, but their hard truncation can occasionally introduce rare erroneous tokens, causing catastrophic errors in long sequences or generate duplicate and overconfident content with low threshold.
\emph{Best-of-$N$ sampling}~\citep{snell2024scaling} generates multiple candidates and selects the best according to a reward model or predefined metric. While effective, it operates at a coarser granularity, is model-dependent, and incurs significant additional computation.
Recently, entropy-based strategies have been explored in large language models (LLMs), where entropy guides model size switch~\citep{simonds2025entropy} or retrieval augmentation~\citep{qiu2025entropybaseddecodingretrievalaugmentedlarge}. 
Similar entropy-guided temperature scaling methods have been proposed for LLMs~\citep{zhang2024edtimprovinglargelanguage} and image generation~\citep{ma2025betterfasterautoregressive} to balance diversity and fidelity.
In contrast to these temperature-scaling approaches, our method leverages entropy to adaptively adjust the Top-$p$ threshold with a $k$-guard mechanism. 
Crucially, we address the video-specific challenge of temporal error accumulation, rather than the static trade-offs focused on in text and image modalities.


\section{Motivation} 



We identify three critical findings in autoregressive (AR) video generation models. These observations explain why static sampling (effective for LLMs) fails for video and therefore motivate our dispersion-aware adaptive sampling strategy. 

\textbf{F1. Video token probability distributions are inherently flat.} 
As shown in Figure~\ref{fig:two_subfigures}, generated video tokens usually have quite small probability values, where dozens of tokens could achieve total probability. 
Tiny gaps between top candidates mean small logit perturbations (model noise, temporal drift) easily flip the argmax, breaking temporal coherence and accumulating visual artifacts. Truncated sampling eases brittleness but remains one-size-fits-all.

The root cause lies in fundamental differences between video and language tokens. Language tokens, with high semantic specificity (each mapping to a clear meaning, e.g., "car") and low redundancy (rarely repeating adjacent tokens). The language tokens have sharp probability distributions where top-1 probabilities often reach 0.7–0.8 (e.g., "street" in "I walk along the \_\_\_"). In contrast, video tokens lack direct semantic grounding  and have high spatiotemporal redundancy. With no single token carrying unique meaning, their distributions are flat and diffuse, with an average top-1 probability of just 0.2.

\begin{figure}[tp]
  \centering
\includegraphics[width=\linewidth]{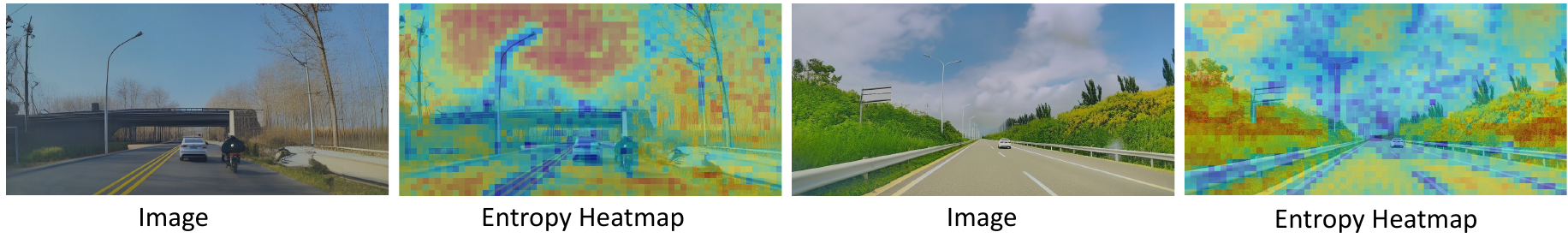}
  \caption{The visualization of entropy heatmaps. High-entropy regions form repeating textures
(e.g., sky, foliage, and road), while low-entropy regions cluster in structured content and distinguishable textures (e.g., boundaries,
edges between sky and trees, road markers and lines).}
  \label{fig:entropy_structure}
  \vspace{-2em}
\end{figure}

\textbf{F2. Video token probability dispersion is inherently tied to the semantic structure of the image}.
Given an AR model, the predicted distribution at the token $i$ is $P_i$, and the entropy is as follows.
\begin{equation}
    H_i=-\sum_{j}^{}p_i(j)logp_i(j)
\end{equation}
where $j$ is the vocabulary index. 

Entropy is a tool to measure the model's uncertainty~\cite{stolfo2024confidence, kang2025scalable} and the token probability dispersions in predictions. Higher entropy indicates low confidence, thus showing greater dispersion in distributions, while lower entropy (high confidence) presents more concentrated distributions. Figure~\ref{fig:entropy_spread} demonstrates how the entropy heatmaps links the generation quality. High-dispersion (high-entropy) regions form repeating textures (e.g., sky, foliage, and road),  where multiple tokens are equally plausible due to subtle texture variations. In contrast, low-dispersion (low-entropy) regions cluster in structured content and distinguishable textures (e.g., boundaries, edges between sky and trees, road markers and lines), where only a few tokens match the stable pixel patterns. However, existing static sampling (top-k or top-p) ignores this structure, forcing large candidate pools on low-dispersion regions (introducing redundant noise) and small pools on high-dispersion ones (discarding valid tokens), thereby exacerbating error accumulation.

\textbf{F3. \emph{Entropy Collapse} in Long-Horizon Autoregressive Video Generation.}  
A third critical finding is that AR video models suffer from entropy collapse during long-horizon generation—an issue tied to evolving token dispersion patterns. As shown in Figure \ref{fig:entropy_spread}, this collapse manifests in two ways: temporally, the share of low-dispersion (low-entropy) tokens grows rapidly with each frame, driving down frame-averaged entropy; spatially, low-dispersion regions expand outward, gradually encroaching on high-dispersion areas, which erodes fine textures (e.g., foliage, road cracks) and replaces them with oversmoothed, uniform blocks (e.g., a detailed tree reduced to solid green). This collapse stems from the model overcommitting to low-dispersion token choices as generation proceeds, which reinforces structural drift and texture wash-out. Notably, entropy collapse is unique to video generation: LLMs avoid it because the high semantic density of language tokens prevents overconfidence in redundant sequences.

\textbf{Insights.}
To address these challenges, we propose a locally adaptive, entropy-guided sampling strategy: align candidate pool size with token dispersion. For low-dispersion regions, small pools (with a minimal 
guard-n) suppress redundant noise and prevent entropy collapse, preserving structural stability while retaining baseline stochasticity. For high-dispersion regions, large pools include all plausible tokens to avoid brittle argmax flips and mitigate early error accumulation. The minimal
guard-n is key—it avoids the extremes of greedy decoding (accelerating texture wash-out) and over-large pools (introducing noise), balancing stability and richness. Details of the entropy-to-k mapping for efficient implementation are provided in Section \ref{sec:method}.

\section{Method}

\label{sec:method}
\begin{figure}[tp]
  \centering
  \includegraphics[width=\linewidth]{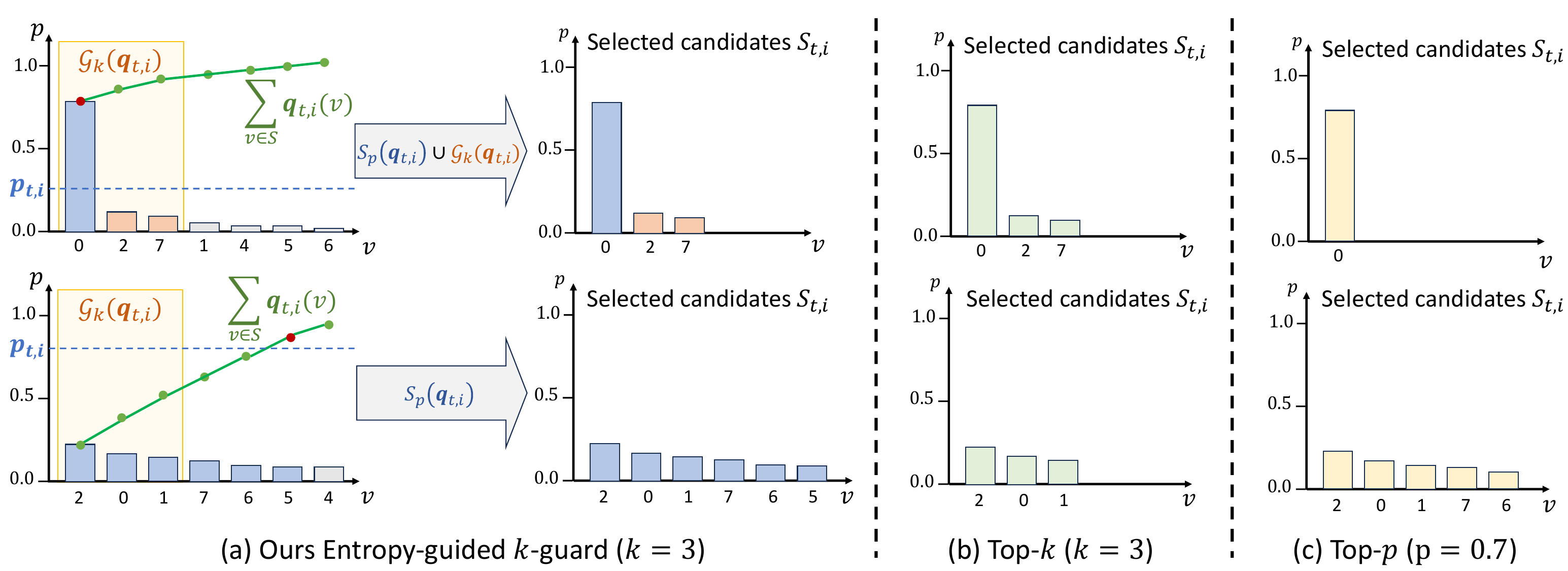}
  \caption{The overall illustration of our sampling strategy.}
  \label{fig:main}
  \vspace{-1 em}
\end{figure}

Inspired by the findings in Section \ref{sec:motivation}, we propose an Uncertainty-aware Adaptive Sampling strategy. The core idea is to leverage the model's predictive uncertainty at each token to dynamically guide the sampling process. Specifically, for regions where the model is confident, we enforce a conservative, near-greedy sampling to preserve structure. Conversely, for ambiguous regions, we encourage more diversity to mitigate brittle decisions and enrich textures. This is implemented through a three-stage process: (1) quantifying token-wise uncertainty using entropy, (2) mapping this uncertainty to an adaptive nucleus threshold, and (3) incorporating a "k-guard" to ensure robust exploration. We provide a pseudocode in Alg. \ref{alg:entropy_mapped_sampling}.

\subsection{Preliminary}



We begin by formalizing the \textbf{autoregressive (AR) formulation} for video generation. Let $\mathcal{V}$ denote a discrete codebook of size $V$, obtained via a learned tokenizer such as VQ-VAE. Each video frame is represented as a grid of tokens $\{z_{t,i} \in \mathcal{V}\}$, where $t$ indexes the temporal step and $i \in \{1, 2, \dots, m\}$ indexes the spatial positions within a frame (each frame contains $m$ VQ tokens).  

An AR world model factorizes the joint distribution of tokens as a product of conditional probabilities. Specifically, the probability of generating the $i$-th token in frame $t$ is conditioned on all previously decoded tokens in the same frame, as well as historical context and actions:

\begin{equation}
p(z_{t,1:i} \mid z_{<t}, c_{<t}, a_{<t})
= \prod_{j=1}^{i} p(z_{t,j} \mid z_{<t}, z_{t,<j}, c_{<t}, a_{<t}).
\end{equation}

where $z_{t,<i}$ denotes the previously decoded tokens within the current frame, $c_{<t}$ represents observed context such as conditioning frames, and $a_{<t}$ denotes historical actions. When $i = m$, the above product gives the joint probability of generating the entire $t$-th frame.  

\begin{figure}[t]
  \centering
  \includegraphics[width=\linewidth]{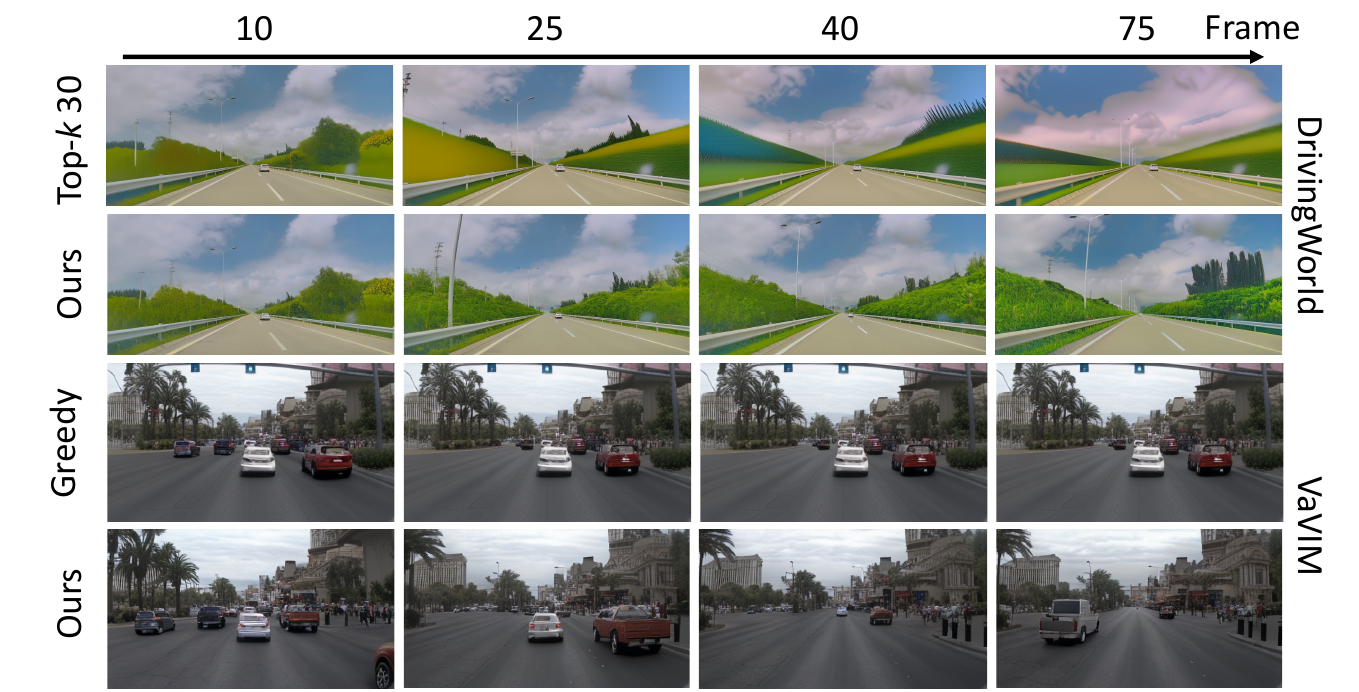}
  \caption{Visual results of DrivingWorld and VaVIM models with our strategy..}
  \label{fig:demo}
  \vspace{-4mm}
\end{figure}

\begin{algorithm}[tp]
\caption{ENkG Sampling}
\label{alg:entropy_mapped_sampling}
\begin{algorithmic}[1]
\REQUIRE probability distribution $\mathbf{p}  \in \mathbb{R}^{ V}$; 
         hyperparameters  $(\alpha, \beta, p_{\text{low}}, p_{\text{high}}, k_{\text{g}})$
         
\ENSURE Sampled token indices $\mathbf{y} \in \mathbb{Z}$

    \STATE Compute normalized entropy:
           $\mathcal{H} \gets - \frac{1}{\log V}\sum_i p_i \log p_i \,$
    \STATE Map entropy to nucleus probability via affine and clip:
       \[
       p_{t,i} \gets \operatorname{clip}\!\big(\alpha\,\widehat{\mathcal{H}}_{t,i} + \beta,\; p_{\mathrm{low}},\; p_{\mathrm{high}}\big)
       \]

        \STATE Let $\{q_{(i)}\}_{i=1}^V$ be probabilities sorted in descending order
        \STATE Find cutoff $c = \min \{ j : \sum_{i=1}^j q_{(i)} \geq p_{t,i} \}$
        \STATE Set $c \gets \max(c,\, k_{\text{g}})$
        \STATE Define truncated distribution 
               $\tilde{q}_i = \frac{q_{(i)}}{\sum_{j=1}^c q_{(j)}} \;\; \text{for } i \leq c,\; 0 \text{ otherwise}$
        \STATE Sample token $y \sim \tilde{q}$

\RETURN $\mathbf{y}$
\end{algorithmic}
\end{algorithm}
\vspace{-1 em}
\subsection{Instability in Autoregressive Video Generation}
However, autoregressive token generation is inherently prone to instability due to \textbf{error accumulation}. Let $\hat{z}_{t,j}$ denote the token actually generated at step $j$. Then the conditional probability for the next token depends on previously generated (potentially erroneous) tokens:

\begin{equation}
    p(z_{t,i} \mid \hat{z}_{t,<i}, c_{<t}, a_{<t}) \neq p(z_{t,i} \mid z_{t,<i}, c_{<t}, a_{<t}),
\end{equation}

where $\hat{z}_{t,<i}$ contains tokens that may differ from the ground truth $z_{t,<i}$. Consequently, small errors propagate through the sequence, amplifying discrepancies in later tokens and potentially degrading entire frames.  

During inference, the model generates a video sequence by sequentially sampling tokens from these categorical distributions. The choice of sampling strategy is therefore critical: greedy decoding often produces blurry frames or repetitive collapse, while excessively random sampling amplifies noise and disrupts temporal coherence. To address this limitation, we introduce an \textbf{Entropy-Guided k-Guard (ENkG) sampling strategy} that dynamically adjusts sampling diversity according to the model’s predictive confidence, balancing structural fidelity and the richness of textures.






\subsection{Entropy-Guided k-Guard sampling}



To quantify token-level uncertainty, we consider the predicted categorical distribution $\boldsymbol{q}_{t,i}$ for each token $z_{t,i}$ at image token site $(t,i)$:
\begin{equation}
q_{t,i}(v) := p\big(z_{t,i} = v \;\big|\; z_{<t},\, z_{t,<i},\, c_{\le t},\, a_{<t}\big), \quad v \in \mathcal{V},
\end{equation}
where $\mathcal{V}$ denotes the discrete codebook. The uncertainty associated with this prediction is measured by its Shannon entropy:
\begin{equation}
\mathcal{H}_{t,i} = -\sum_{v \in \mathcal{V}} q_{t,i}(v) \log q_{t,i}(v).
\end{equation}
To obtain a standardized measure on the unit interval, we normalize the entropy by the maximum possible value $\log|\mathcal{V}|$:
\begin{equation}
\widehat{\mathcal{H}}_{t,i} = \frac{\mathcal{H}_{t,i}}{\log |\mathcal{V}|} \in [0,1].
\end{equation}

The normalized entropy $\widehat{\mathcal{H}}_{t,i}$ serves as a direct indicator of the model's confidence in predicting token $z_{t,i}$. Low values of $\widehat{\mathcal{H}}_{t,i}$ correspond to sharply peaked distributions, indicating high confidence in a dominant token, whereas high values indicate flatter distributions and thus greater uncertainty. Specifically, $\widehat{\mathcal{H}}_{t,i} \approx 0$ corresponds to a highly confident, nearly deterministic prediction, while $\widehat{\mathcal{H}}_{t,i} \approx 1$ corresponds to a nearly uniform, highly uncertain prediction.

\paragraph{Entropy-guided adaptive nucleus.} 
To adaptively control sampling diversity, the normalized predictive entropy $\widehat{\mathcal{H}}_{t,i}$ is mapped to a target cumulative probability $p_{t,i} \in [p_{\mathrm{low}}, p_{\mathrm{high}}]$ via an affine transformation with clipping:
\begin{equation}
\begin{aligned}
    p&_{t,i} = \mathrm{clip}\Big(\alpha \,\widehat{\mathcal{H}}_{t,i} + \beta, \; p_{\mathrm{low}}, \; p_{\mathrm{high}}\Big). \\
\textbf{where }& \alpha = \frac{p_{\mathrm{high}} - p_{\mathrm{low}}}{\widehat{\mathcal{H}}_{\mathrm{high}} - \widehat{\mathcal{H}}_{\mathrm{low}}},
\beta = p_{\mathrm{low}} - \alpha \,\widehat{\mathcal{H}}_{\mathrm{low}}.
\end{aligned}
\end{equation}
\label{eq:alpha-beta}
where $\mathrm{clip}(x,a,b) = \min(\max(x,a),b)$. In the experiments, $p_{\mathrm{low}} = 0.65$, $p_{\mathrm{high}} = 0.9$, and $\widehat{\mathcal{H}}_{\mathrm{low}} = 0.25$, $\widehat{\mathcal{H}}_{\mathrm{high}} = 0.6$. 

Based on $p_{t,i}$, the adaptive nucleus set $\mathcal{S}_p(\boldsymbol{q}_{t,i})$ is defined as the minimal subset of tokens whose cumulative probability meets or exceeds $p_{t,i}$:
\begin{equation}
\mathcal{S}_{p}(\boldsymbol{q}_{t,i}) = \arg\min_{\mathcal{S} \subseteq \mathcal{V}} 
\Big\{ |\mathcal{S}| \;\Big|\; \sum_{v \in \mathcal{S}} q_{t,i}(v) \ge p_{t,i} \Big\}.
\end{equation}

Tokens with low entropy correspond to high-confidence predictions, allowing a small nucleus and near-greedy sampling that preserves fine structures such as edges and boundaries. High-entropy tokens indicate greater uncertainty; a larger nucleus in these cases encourages exploration, enhances diversity, and mitigates the compounding of errors in sequential autoregressive generation. This entropy-guided adaptation provides a principled mechanism to balance fidelity and diversity in token-level sampling.

\paragraph{$k$-Guard for robust exploration.} 
Direct entropy-guided adaptive sampling can become overly greedy in high-confidence (low-entropy) regions, where the nucleus set may contain only a few tokens. To preserve minimal exploration without compromising stability, the nucleus is augmented with the $k_g$ most probable tokens, forming a $k$-guard:
\begin{equation}
\mathcal{S}_{t,i} = \mathcal{S}_p(\boldsymbol{q}_{t,i}) \;\cup\; \mathcal{G}_{k}(\boldsymbol{q}_{t,i}, k_g),
\end{equation}
where $\mathcal{S}_p(\boldsymbol{q}_{t,i})$ denotes the nucleus (top-$p$) candidate set, and $\mathcal{G}_{k}(\boldsymbol{q}_{t,i}, k_g)$ returns the indices of the $k_g$ tokens with the highest probabilities under $\boldsymbol{q}_{t,i}$. Typical choices for $k_g$ are small integers such as $3$, $5$, or $10$. If the nucleus already contains these tokens, the union leaves the set unchanged. To limit computational cost, a maximum size $n_{\max}$ can be enforced by retaining only the top $n_{\max}$ tokens sorted by probability.

Token selection is then performed by sampling from the renormalized distribution over $\mathcal{S}_{t,i}$. This combined framework leverages token-wise uncertainty to adaptively regulate sampling diversity, while the $k$-guard ensures minimal exploration in highly confident regions, enhancing the robustness and stability of sequential autoregressive generation.

\begin{table}[tp]
\caption{Quantitative results on DiverseDrive and Nuplan. \textit{* Cosmos uses a fixed 33-frame generation window; hence its metrics are computed on the first 33 frames (vs.\ 75 for others).}}
\label{tab:main}
\centering
\resizebox{\textwidth}{!}{%
\begin{tabular}{l|ccccc|ccccc}
\hline
\multirow{2}{*}{Model}
& \multicolumn{5}{c|}{DiverseDrive} 
& \multicolumn{5}{c}{Nuplan} \\
&$\text{FVD}_{75}$$\downarrow$ & $\text{FID}_{75}$$\downarrow$ & LPIPS$\downarrow$ & PSNR$\uparrow$ & SSIM $\uparrow$
&$\text{FVD}_{75}$$\downarrow$ & $\text{FID}_{75}$$\downarrow$ & LPIPS$\downarrow$ & PSNR$\uparrow$ & SSIM $\uparrow$ \\
\hline
DrivingWorld(top-$k$~30)
  & 696 & 61.78 & 0.401 & 14.03 & 0.43          & 583    & 37.80 & 0.380 & 14.22 & 0.39 \\
DrivingWorld(+Ours)    
  & \textbf{489} & \textbf{26.61} & \textbf{0.350} & \textbf{15.87} & \textbf{0.45}          & \textbf{565}    & \textbf{31.34} & \textbf{0.360} & \textbf{14.96} & \textbf{0.40} \\
\hline
VaVIM(greedy)
  & 1473   & 91.75 & 0.396 & 16.46 & 0.50  & {927}    & {65.26} & {0.315} & {14.82} & {0.44} \\

VaVIM(+Ours)    
  & \textbf{1055}   & \textbf{46.76} & {0.426} & 14.76 & 0.46   & 1031   & \textbf{41.60} & 0.327 & 14.43 & 0.42 \\
\hline
Cosmos(top-$p$~0.8)*
   & 1260 & 87.82  & 0.48   & 16.56 &0.54  
    & 814    &80.45   & 0.29   &17.52   & 0.54 \\
Cosmos(+Ours)*
 & \textbf{1132} & \textbf{84.67}  & \textbf{0.47}  & \textbf{16.61} &0.53     
 & \textbf{801} & \textbf{75.01}   & 0.29  & \textbf{17.81} & \textbf{0.55}  \\
\hline

\end{tabular}
}
\vspace{-2 em}
\end{table}

\section{Experiments}
\subsection{Experimental Setting}
\textbf{Models.} Since ENkG is a plug-and-play solution, we integrate it with existing AR-based video world model for experiments, including DrivingWorld~\citep{hu2024drivingworld}, VaVIM~\citep{vavam2025} and Cosmos~\citep{nvidia2025cosmos}. We keep generation parameters consistent with the original model for fair  comparison. Specifically, DrivingWorld adopts top-$k$($k=30$), VaVIM adopts greedy sampling, while Cosmos uses a top-$p$ ($p=0.8$).  

\textbf{Evaluation Dataset. }
We conduct evaluations on two datasets: DiverseDrive and nuPlan. 
DiverseDrive is a self-collected high-quality driving dataset, which consists of 50 video clips. Compared to nuPlan datasets, DiverseDrive contains more scenarios and a richer variety of plants.
These characteristics promote stronger generalization, making DiverseDrive a closer match to the requirements of world-model evaluation.

\begin{figure}[tp]
  \centering
  \includegraphics[width=0.9\linewidth]{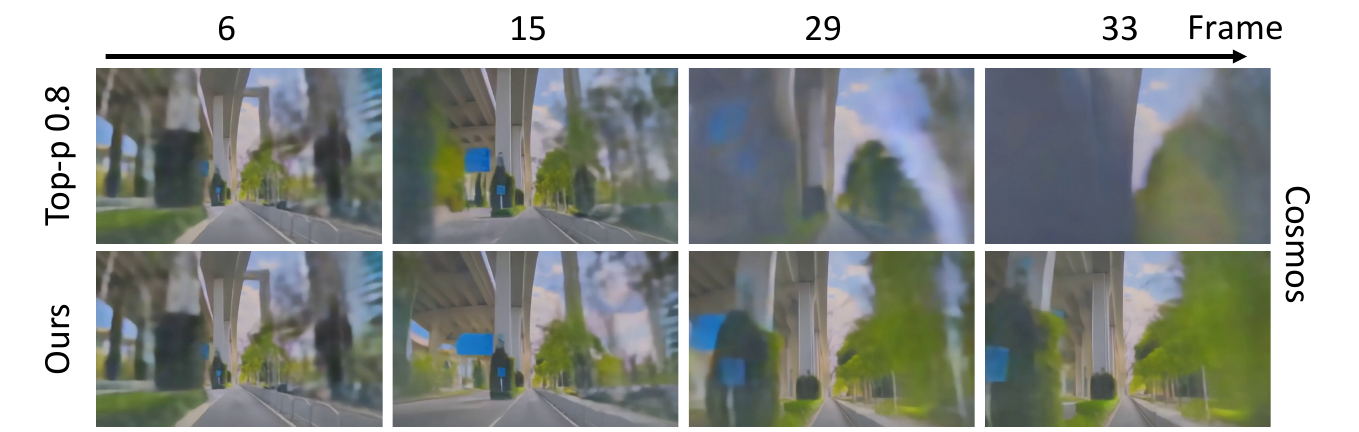}
  \caption{Visual results of Cosmos model with our strategy.}
  \label{fig:demo2}
  \vspace{-4mm}
\end{figure}

\textbf{Metrics.} To assess the quality of generated videos, we report the Fréchet Video Distance (FVD) as a measure of video-level realism, and the Fréchet Inception Distance (FID) to evaluate per-frame image fidelity. 
In addition, we include low-level metrics such as LPIPS, PSNR, and SSIM as supplementary evaluations, though these metrics are not well-suited for the video generation task.
\subsection{Main Results}

\textbf{Quantitative Comparison.} 
As shown in Table~\ref{tab:main}, integrating ENkG consistently yields substantial gains across different architectures. On DiverseDrive, our method reduces FVD and FID by an average of 22.8\% and 36.5\% respectively, while also lowering LPIPS and improving PSNR/SSIM, indicating both perceptual and structural benefits. DrivingWorld also benefits from ENkG on Nuplan, despite being sufficiently trained on this dataset. Even Cosmos, which has a relatively weak AR backbone, achieves modest improvements on DiverseDrive. Notably, VaVIM tends to generate repeated frames, which artificially yields relatively lower FVD values; our strategy, by contrast, effectively alleviates this frame-freezing issue.





\begin{table}[tbp] 
  \centering
  \setlength{\tabcolsep}{3pt} 
  \parbox[t]{0.8\linewidth}{ 
    \centering
    \small
    \caption{Ablation study with DrivingWorld model on self-collected dataset.}
    \begin{tabular}{lcccc} 
    \hline
    Method    &    FVD$\downarrow$ & FID $\downarrow$ & LPIPS$\downarrow$ & PSNR$\uparrow$ \\
    \hline
    Full Strategy & \textbf{489} & \textbf{26.61} & \textbf{0.350} & \textbf{15.87} \\
    w/o Entropy   & 532 & 41.43 & 0.591 & 13.96 \\
    w/o k-Guard   & 552 & 39.76 & 0.421 & 15.18 \\
    \hline
    \end{tabular}
  }
  \vspace{-8mm}
\end{table}




\textbf{Qualitative Comparison.} As shown in Fig.~\ref{fig:demo} and Fig.~\ref{fig:demo2}, the existing sampling techniques  frequently result in textural degradation, where crucial details in road markings, such as crosswalks, and surrounding vegetation become indistinct and blurry. 
Furthermore, these approaches are susceptible to color distortion, leading to unnatural and washed-out hues that compromise the scene's realism. 
The generated sequence of VaVIM collapses into a static or near-static frame, failing to capture the inherent dynamics of the driving environment and rendering vehicles motionless. 
In contrast, our entropy-guided approach, which dynamically adjusts the size of candidates to prevent overconfidence, demonstrates substantial improvements in perceptual quality. 
Our strategy effectively mitigates the aforementioned issues, producing videos with sharp, well-defined textures and accurate color fidelity. 

\subsection{Ablation Study}

\begin{figure}[tp]
    \centering
    \includegraphics[width=\linewidth]{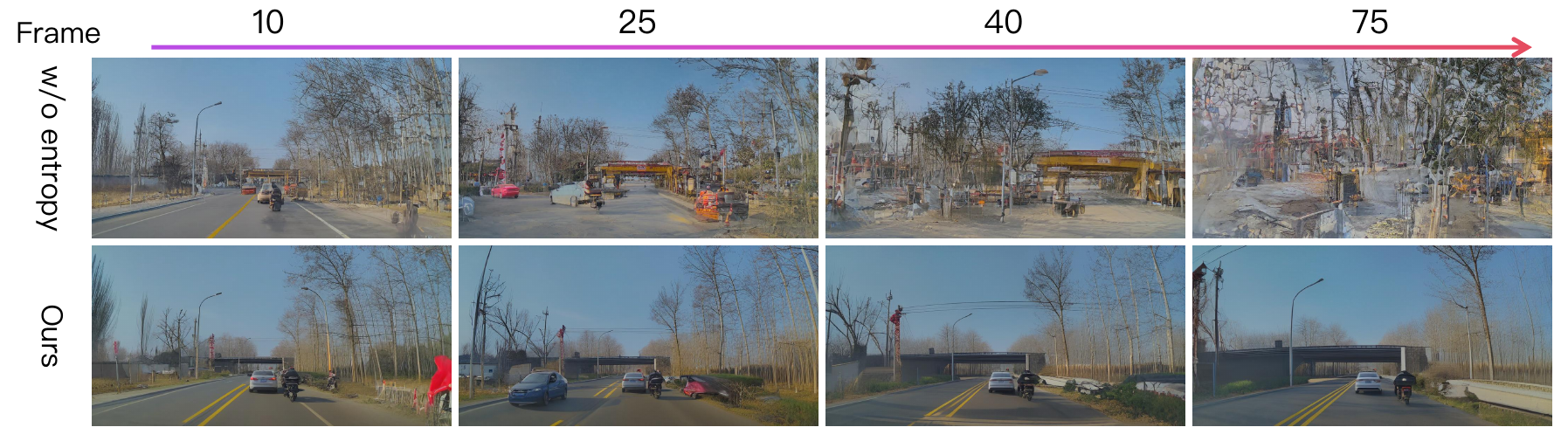}
    \caption{Entropy-adaptive guidance prevents collapse in the video model.}
    \label{fig:ablation_entropy}
    \vspace{-1em}
\end{figure}
\begin{figure}
    \centering
    \includegraphics[width=\linewidth]{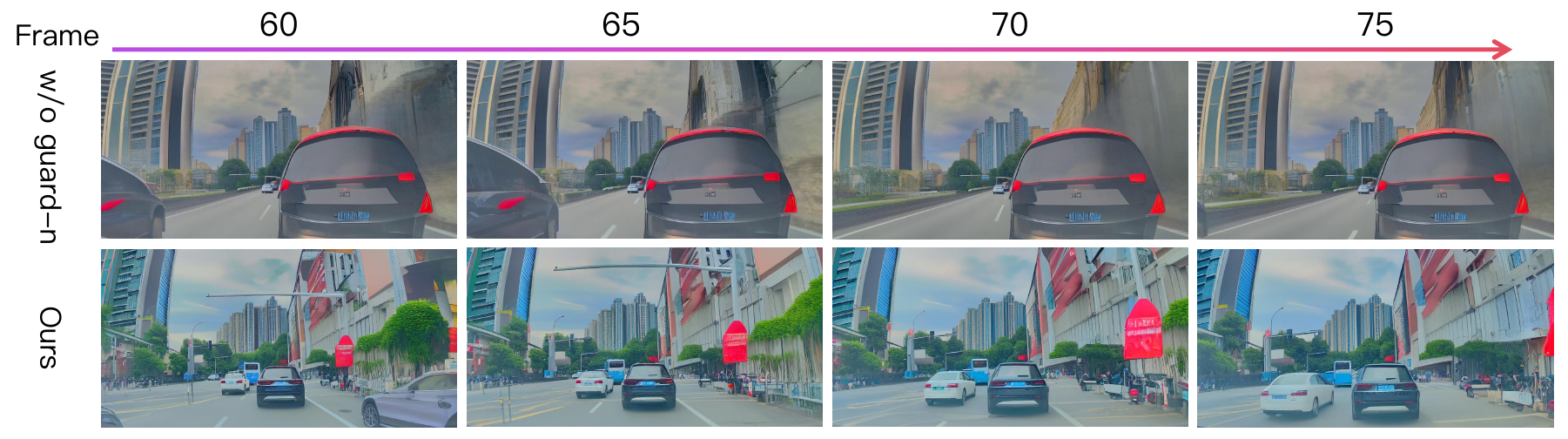}
    \caption{The k-guard design prevents frame-freezing in the video model.}
    \label{fig:ablation_kguard}
    \vspace{-2em}
\end{figure}

\textbf{Effect of entropy-adaptive guidance.}
The core contribution of the entropy-adaptive guidance is the dynamic adjustment of the sampling nucleus based on the model's predictive uncertainty. As shown in Figure \ref{fig:entropy_spread} and Figure \ref{fig:ablation_entropy}, it effectively mitigates issues of textural decay and color shifting commonly seen in baseline methods. By allowing a wider range of tokens when uncertainty is high and narrowing it when the model is confident, our method preserves high-frequency details, resulting in significantly sharper textures on surfaces like road markings and vegetation. This enhancement in per-frame visual fidelity translates to lower FID and LPIPS scores, indicative of more realistic and perceptually similar generated frames. Consequently, entropy-adaptive guidance significantly improves the visual quality and realism of the generated videos.

\textbf{Effect of $k$-guard.}
The $k$-guard mechanism ensures a minimum level of diversity in candidate tokens. Without the $k$-guard, the model can, even with entropy-adaptive guidance, become overly confident in certain contexts. In Figure~\ref{fig:ablation_kguard}, this leads to significant temporal artifacts, such as vehicles that remain nearly stationary when they should be in motion, a physically implausible scenario. The introduction of $k$-guard directly addresses this failure mode, leading to more fluid and realistic motion dynamics, as quantitatively reflected in the substantial reduction of the FVD score, which measures temporal consistency. Therefore, the $k$-guard is crucial for maintaining temporal coherence and preventing the generation of degenerate, static sequences.


\section{Conclusion}
\label{sec:Conclusion}

In this work, we investigated the challenge of error accumulation in autoregressive video generation and highlighted the overlooked role of the sampling process. We proposed \emph{Uncertainty-aware Adaptive Sampling}, a simple yet effective strategy that dynamically modulates token diversity based on predictive entropy with a minimal $k$-guard. Unlike prior approaches that require architectural changes or retraining, our method operates purely at inference time, making it broadly applicable to existing large-scale models. Extensive experiments demonstrate that our approach significantly improves temporal coherence, preserves fine-grained details, and extends the effective generation horizon. These results suggest that inference-time uncertainty-aware strategies provide a practical and generalizable path toward more robust and high-fidelity video world models.

\bibliography{main}

@article{hu2024drivingworld,
  title={DrivingWorld: ConstructingWorld Model for Autonomous Driving via Video GPT},
  author={Hu, Xiaotao and Yin, Wei and Jia, Mingkai and Deng, Junyuan and Guo, Xiaoyang and Zhang, Qian and Long, Xiaoxiao and Tan, Ping},
  journal={arXiv preprint arXiv:2412.19505},
  year={2024}
}

@article{nvidia2025cosmos,
  title   = {Cosmos World Foundation Model Platform for Physical AI},
  author  = {NVIDIA},
  journal = {arXiv preprint arXiv:2501.03575},
  year    = {2025},
  url     = {https://arxiv.org/abs/2501.03575}
}

@article{vavam2025,
  title={VaViM and VaVAM: Autonomous Driving through Video Generative Modeling},
  author={Bartoccioni, Florent and Ramzi, Elias and Besnier, Victor and Venkataramanan, Shashanka and Vu, Tuan-Hung and Xu, Yihong and Chambon, Loick and Gidaris, Spyros and Odabas, Serkan and Hurych, David and Marlet, Renaud and Boulch, Alexandre and Chen, Mickael and Zablocki, Eloi and Bursuc, Andrei and Valle, Eduardo and Cord, Matthieu},
  journal={arXiv preprint arXiv:2502.15672},
  year={2025}
}

@misc{openai2024video,
  author={OpenAI},
  title={Video generation models as world simulators},
  year={2024},
  howpublished={\url{https://openai.com/research/video-generation-models-as-world-simulators}}
}

@article{schmidt2019generalization,
  title={Generalization in Generation: A closer look at Exposure Bias},
  author={Schmidt, Florian},
  journal={EMNLP-IJCNLP 2019},
  pages={157},
  year={2019}
}

@article{dosovitskiy2020image,
  title={An image is worth 16x16 words: Transformers for image recognition at scale},
  author={Dosovitskiy, Alexey and Beyer, Lucas and Kolesnikov, Alexander and Weissenborn, Dirk and Zhai, Xiaohua and Unterthiner, Thomas and Dehghani, Mostafa and Minderer, Matthias and Heigold, Georg and Gelly, Sylvain and others},
  journal={arXiv preprint arXiv:2010.11929},
  year={2020}
}

@misc{weissenbornscaling,
      title={Scaling Autoregressive Video Models}, 
      author={Dirk Weissenborn and Oscar Täckström and Jakob Uszkoreit},
      year={2020},
      eprint={1906.02634},
      archivePrefix={arXiv},
      primaryClass={cs.CV},
      url={https://arxiv.org/abs/1906.02634}, 
}

@article{snell2024scaling,
  title={Scaling llm test-time compute optimally can be more effective than scaling model parameters},
  author={Snell, Charlie and Lee, Jaehoon and Xu, Kelvin and Kumar, Aviral},
  journal={arXiv preprint arXiv:2408.03314},
  year={2024}
}

@inproceedings{ravfogel2023conformal,
  title={Conformal Nucleus Sampling},
  author={Ravfogel, Shauli and Goldberg, Yoav and Goldberger, Jacob},
  booktitle={Findings of the Association for Computational Linguistics: ACL 2023},
  pages={27--34},
  year={2023}
}

@misc{noarov2025foundationstopkdecodinglanguage,
      title={Foundations of Top-$k$ Decoding For Language Models}, 
      author={Georgy Noarov and Soham Mallick and Tao Wang and Sunay Joshi and Yan Sun and Yangxinyu Xie and Mengxin Yu and Edgar Dobriban},
      year={2025},
      eprint={2505.19371},
      archivePrefix={arXiv},
      primaryClass={cs.AI},
      url={https://arxiv.org/abs/2505.19371}, 
}

@article{ha2018world,
  title={World Models},
  author={Ha, David and Schmidhuber, J{\"u}rgen},
  journal={CoRR},
  year={2018}
}

@article{ding2024understanding,
  title={Understanding world or predicting future? a comprehensive survey of world models},
  author={Ding, Jingtao and Zhang, Yunke and Shang, Yu and Zhang, Yuheng and Zong, Zefang and Feng, Jie and Yuan, Yuan and Su, Hongyuan and Li, Nian and Sukiennik, Nicholas and others},
  journal={ACM Computing Surveys},
  year={2024},
  publisher={ACM New York, NY}
}

@misc{long2025surveylearningembodiedintelligence,
      title={A Survey: Learning Embodied Intelligence from Physical Simulators and World Models}, 
      author={Xiaoxiao Long and Qingrui Zhao and Kaiwen Zhang and Zihao Zhang and Dingrui Wang and Yumeng Liu and Zhengjie Shu and Yi Lu and Shouzheng Wang and Xinzhe Wei and Wei Li and Wei Yin and Yao Yao and Jia Pan and Qiu Shen and Ruigang Yang and Xun Cao and Qionghai Dai},
      year={2025},
      eprint={2507.00917},
      archivePrefix={arXiv},
      primaryClass={cs.RO},
      url={https://arxiv.org/abs/2507.00917}, 
}

@article{parthipan2024defining,
  title={{Defining error accumulation in ML atmospheric simulators}},
  author={Parthipan, Raghul and Anand, Mohit and Christensen, Hannah M and Hosking, J Scott},
  journal={arXiv preprint arXiv:2405.14714},
  year={2024}
}

@article{bengio2015scheduled,
  title={{Scheduled Sampling for Sequence Prediction with Recurrent Neural Networks}},
  author={Bengio, Samy and Vinyals, Oriol and Jaitly, Navdeep and Shazeer, Noam},
  journal={arXiv preprint arXiv:1506.03099},
  year={2015}
}

@article{hu2023gaia,
  title={Gaia-1: A generative world model for autonomous driving},
  author={Hu, Anthony and Russell, Lloyd and Yeo, Hudson and Murez, Zak and Fedoseev, George and Kendall, Alex and Shotton, Jamie and Corrado, Gianluca},
  journal={arXiv preprint arXiv:2309.17080},
  year={2023}
}

@misc{ross2011reductionimitationlearningstructured,
      title={A Reduction of Imitation Learning and Structured Prediction to No-Regret Online Learning}, 
      author={Stephane Ross and Geoffrey J. Gordon and J. Andrew Bagnell},
      year={2011},
      eprint={1011.0686},
      archivePrefix={arXiv},
      primaryClass={cs.LG},
      url={https://arxiv.org/abs/1011.0686}, 
}

@misc{ren2025xar,
      title={Beyond Next-Token: Next-X Prediction for Autoregressive Visual Generation}, 
      author={Sucheng Ren and Qihang Yu and Ju He and Xiaohui Shen and Alan Yuille and Liang-Chieh Chen},
      year={2025},
      eprint={2502.20388},
      archivePrefix={arXiv},
      primaryClass={cs.CV},
      url={https://arxiv.org/abs/2502.20388}, 
}

@misc{zhou2025magi,
      title={Taming Teacher Forcing for Masked Autoregressive Video Generation}, 
      author={Deyu Zhou and Quan Sun and Yuang Peng and Kun Yan and Runpei Dong and Duomin Wang and Zheng Ge and Nan Duan and Xiangyu Zhang and Lionel M. Ni and Heung-Yeung Shum},
      year={2025},
      eprint={2501.12389},
      archivePrefix={arXiv},
      primaryClass={cs.CV},
      url={https://arxiv.org/abs/2501.12389}, 
}

@article{simonds2025entropy,
  title={Entropy Adaptive Decoding: Dynamic Model Switching for Efficient Inference},
  author={Simonds, Toby},
  journal={arXiv preprint arXiv:2502.06833},
  year={2025}
}

@misc{qiu2025entropybaseddecodingretrievalaugmentedlarge,
      title={Entropy-Based Decoding for Retrieval-Augmented Large Language Models}, 
      author={Zexuan Qiu and Zijing Ou and Bin Wu and Jingjing Li and Aiwei Liu and Irwin King},
      year={2025},
      eprint={2406.17519},
      archivePrefix={arXiv},
      primaryClass={cs.CL},
      url={https://arxiv.org/abs/2406.17519}, 
}

@article{wu2024ivideogpt,
  title={ivideogpt: Interactive videogpts are scalable world models},
  author={Wu, Jialong and Yin, Shaofeng and Feng, Ningya and He, Xu and Li, Dong and Hao, Jianye and Long, Mingsheng},
  journal={Advances in Neural Information Processing Systems},
  volume={37},
  pages={68082--68119},
  year={2024}
}

@article{gu2025long,
  title={Long-context autoregressive video modeling with next-frame prediction},
  author={Gu, Yuchao and Mao, Weijia and Shou, Mike Zheng},
  journal={arXiv preprint arXiv:2503.19325},
  year={2025}
}

@misc{li2025roboticworldmodelneural,
      title={Robotic World Model: A Neural Network Simulator for Robust Policy Optimization in Robotics}, 
      author={Chenhao Li and Andreas Krause and Marco Hutter},
      year={2025},
      eprint={2501.10100},
      archivePrefix={arXiv},
      primaryClass={cs.RO},
      url={https://arxiv.org/abs/2501.10100}, 
}

@article{ho2022video,
title={Video diffusion models},
author={Ho, Jonathan and Salimans, Tim and Gritsenko, Alexey and Chan, William and Norouzi, Mohammad and Fleet, David J},
journal={arXiv:2204.03458},
year={2022}}

@article{wang2023worlddreamer,
  title={WorldDreamer: Towards General World Models for Video Generation via Predicting Masked Tokens},
  author={Wang, Xiaofeng and Zhu, Zheng and Huang, Guan and Wang, Boyuan and Chen, Xinze and Lu, Jiwen},
  journal={arXiv preprint arXiv:2401.09985},
  year={2024}
}

@article{he2025pretrained,
  title={Pre-Trained Video Generative Models as World Simulators},
  author={He, Haoran and Zhang, Yang and Lin, Liang and Xu, Zhongwen and Pan, Ling},
  journal={arXiv preprint arXiv:2502.07825},
  year={2025}
}

@article{mo2025dreamland,
  title={Dreamland: Controllable World Creation with Simulator and Generative Models},
  author={Mo, Sicheng and Leng, Ziyang and Liu, Leon and Wang, Weizhen and He, Honglin and Zhou, Bolei},
  journal={arXiv preprint arXiv:2506.08006},
  year={2025}
}

@article{villegas2022phenaki,
  title={Phenaki: Variable Length Video Generation From Open Domain Textual Description},
  author={Villegas, Ruben and Babaeizadeh, Mohammad and Kindermans, Pieter‐Jan and Moraldo, Hernan and Zhang, Han and Saffar, Mohammad Taghi and Castro, Santiago and Kunze, Julius and Erhan, Dumitru},
  journal={arXiv preprint arXiv:2210.02399},
  year={2022}
}

@article{kong20253d,
  title={3D and 4D world modeling: A survey},
  author={Kong, Lingdong and Yang, Wesley and Mei, Jianbiao and Liu, Youquan and Liang, Ao and Zhu, Dekai and Lu, Dongyue and Yin, Wei and Hu, Xiaotao and Jia, Mingkai and others},
  journal={arXiv preprint arXiv:2509.07996},
  year={2025}
}

@article{kang2025scalable,
  title={Scalable best-of-n selection for large language models via self-certainty},
  author={Kang, Zhewei and Zhao, Xuandong and Song, Dawn},
  journal={arXiv preprint arXiv:2502.18581},
  year={2025}
}

@article{stolfo2024confidence,
  title={Confidence regulation neurons in language models},
  author={Stolfo, Alessandro and Wu, Ben and Gurnee, Wes and Belinkov, Yonatan and Song, Xingyi and Sachan, Mrinmaya and Nanda, Neel},
  journal={Advances in Neural Information Processing Systems},
  volume={37},
  pages={125019--125049},
  year={2024}
}

@article{Qwen2.5-VL,
  title={Qwen2.5-VL Technical Report},
  author={Bai, Shuai and Chen, Keqin and Liu, Xuejing and Wang, Jialin and Ge, Wenbin and Song, Sibo and Dang, Kai and Wang, Peng and Wang, Shijie and Tang, Jun and Zhong, Humen and Zhu, Yuanzhi and Yang, Mingkun and Li, Zhaohai and Wan, Jianqiang and Wang, Pengfei and Ding, Wei and Fu, Zheren and Xu, Yiheng and Ye, Jiabo and Zhang, Xi and Xie, Tianbao and Cheng, Zesen and Zhang, Hang and Yang, Zhibo and Xu, Haiyang and Lin, Junyang},
  journal={arXiv preprint arXiv:2502.13923},
  year={2025}
}

@article{zhang2025epona,
  title={Epona: Autoregressive Diffusion World Model for Autonomous Driving},
  author={Zhang, Kaiwen and Tang, Zhenyu and Hu, Xiaotao and Pan, Xingang and Guo, Xiaoyang and Liu, Yuan and Huang, Jingwei and Yuan, Li and Zhang, Qian and Long, Xiao-Xiao and others},
  journal={arXiv preprint arXiv:2506.24113},
  year={2025}
}

@misc{zhang2024edtimprovinglargelanguage,
      title={EDT: Improving Large Language Models' Generation by Entropy-based Dynamic Temperature Sampling}, 
      author={Shimao Zhang and Yu Bao and Shujian Huang},
      year={2024},
      eprint={2403.14541},
      archivePrefix={arXiv},
      primaryClass={cs.CL},
      url={https://arxiv.org/abs/2403.14541}, 
}

@misc{huang2025selfforcingbridgingtraintest,
      title={Self Forcing: Bridging the Train-Test Gap in Autoregressive Video Diffusion}, 
      author={Xun Huang and Zhengqi Li and Guande He and Mingyuan Zhou and Eli Shechtman},
      year={2025},
      eprint={2506.08009},
      archivePrefix={arXiv},
      primaryClass={cs.CV},
      url={https://arxiv.org/abs/2506.08009}, 
}

@misc{wang2025erroranalysesautoregressivevideo,
      title={Error Analyses of Auto-Regressive Video Diffusion Models: A Unified Framework}, 
      author={Jing Wang and Fengzhuo Zhang and Xiaoli Li and Vincent Y. F. Tan and Tianyu Pang and Chao Du and Aixin Sun and Zhuoran Yang},
      year={2025},
      eprint={2503.10704},
      archivePrefix={arXiv},
      primaryClass={cs.CV},
      url={https://arxiv.org/abs/2503.10704}, 
}

@misc{ma2025betterfasterautoregressive,
      title={Towards Better \& Faster Autoregressive Image Generation: From the Perspective of Entropy}, 
      author={Xiaoxiao Ma and Feng Zhao and Pengyang Ling and Haibo Qiu and Zhixiang Wei and Hu Yu and Jie Huang and Zhixiong Zeng and Lin Ma},
      year={2025},
      eprint={2510.09012},
      archivePrefix={arXiv},
      primaryClass={cs.CV},
      url={https://arxiv.org/abs/2510.09012}, 
}

@misc{yuan2025lumos1autoregressivevideogeneration,
      title={Lumos-1: On Autoregressive Video Generation from a Unified Model Perspective}, 
      author={Hangjie Yuan and Weihua Chen and Jun Cen and Hu Yu and Jingyun Liang and Shuning Chang and Zhihui Lin and Tao Feng and Pengwei Liu and Jiazheng Xing and Hao Luo and Jiasheng Tang and Fan Wang and Yi Yang},
      year={2025},
      eprint={2507.08801},
      archivePrefix={arXiv},
      primaryClass={cs.CV},
      url={https://arxiv.org/abs/2507.08801}, 
}

@misc{ren2025blockpredictionvideogeneration,
      title={Next Block Prediction: Video Generation via Semi-Autoregressive Modeling}, 
      author={Shuhuai Ren and Shuming Ma and Xu Sun and Furu Wei},
      year={2025},
      eprint={2502.07737},
      archivePrefix={arXiv},
      primaryClass={cs.CV},
      url={https://arxiv.org/abs/2502.07737}, 
}
\bibliographystyle{iclr2026_conference}

\clearpage
\appendix
\section{Appendix}
\subsection{Experimental Protocols }
\label{app:proto}
We report FVD, FID, LPIPS, PSNR, and SSIM under identical preprocessing and frame sampling across methods unless noted. 
\textbf{Cosmos note:} Cosmos uses a fixed 33-frame generation window; thus all Cosmos metrics are computed on the first 33 frames of each sequence, while others use 75 (\(\text{FVD}_{75}\), \(\text{FID}_{75}\)).


\section{Hyperparameter Sensitivity and Baseline Tuning}
\label{app:hyper}

This section provides additional analyses on the sensitivity of ENkG to its hyperparameters and on the tuning of static sampling baselines. Unless otherwise noted, all experiments are conducted on DrivingWorld.

\subsection{Sensitivity of ENkG Hyperparameters}

\paragraph{Robustness to entropy and probability thresholds.}
We first study the robustness of ENkG with respect to the entropy thresholds $(H_{\text{low}}, H_{\text{high}})$ and the corresponding probability range $(p_{\text{low}}, p_{\text{high}})$ used for entropy-guided truncation.
Starting from the default configuration (\emph{Mid}), we construct two extreme variants: a conservative configuration (\emph{Left}), which prefers smaller $p$ and narrower entropy range, and an aggressive configuration (\emph{Right}), which allows larger $p$ and a wider entropy range.

Table~\ref{tab:enk_sensitivity_thresholds} shows that all metrics remain stable across these settings.
Although the extreme variants cause moderate degradation in FVD/FID, the overall differences are small, indicating a broad performance plateau.
This suggests that ENkG does not require delicate hand-tuning of entropy or probability thresholds to remain effective.

\begin{table}[h]
    \centering
    \caption{Sensitivity to entropy thresholds $(H_{\text{low}}, H_{\text{high}})$ and probability band $(p_{\text{low}}, p_{\text{high}})$.
    The default configuration (\emph{Mid}) achieves the best trade-off, while both conservative (\emph{Left}) and aggressive (\emph{Right}) shifts only moderately affect the performance.}
    \label{tab:enk_sensitivity_thresholds}
    \vspace{0.3em}
    \begin{tabular}{lccccccc}
        \toprule
        Setting & $H_{\text{low/high}}$ & $p_{\text{low/high}}$ &
        FVD $\downarrow$ & FID $\downarrow$ &
        LPIPS $\downarrow$ & SSIM $\uparrow$ & PSNR $\uparrow$ \\
        \midrule
        Left  & 0.0 / 0.5  & 0.60 / 0.90 & 522.05 & 30.93 & 0.35 & 0.49 & \textbf{16.26} \\
        \textbf{Mid (Default)} &
        \textbf{0.25 / 0.60} & \textbf{0.65 / 0.90} &
        \textbf{489.00} & \textbf{26.61} & \textbf{0.35} & \textbf{0.45} & 15.87 \\
        Right & 0.40 / 0.90 & 0.80 / 0.95 & 497.52 & 29.91 & 0.35 & 0.48 & 15.98 \\
        \bottomrule
    \end{tabular}
\end{table}

\paragraph{Insensitivity to guard size $k_g$.}
Next, we fix all other ENkG hyperparameters and vary the guard size $k_g$, which controls how many top candidates are preserved by the $k$-guard at each step.
As shown in Table~\ref{tab:enk_sensitivity_kg}, the performance is highly stable for any $k_g \in [2, 15]$, and only the degenerate case $k_g=1$ (which effectively disables the guard) leads to a clear degradation.
This indicates that while the \emph{presence} of the $k$-guard is crucial to prevent collapse, its exact value is not sensitive in a wide range.

\begin{table}[t]
    \centering
    \caption{Sensitivity to guard size $k_g$ (other hyperparameters fixed).
    ENkG remains stable for a broad range of $k_g$, and only the degenerate case $k_g=1$ (no guard) leads to noticeable degradation.}
    \label{tab:enk_sensitivity_kg}
    \vspace{0.3em}
    \begin{tabular}{l|ccccc}
        \toprule
        $k_g$ & 1 & 2 & \textbf{3 (Default)} & 7 & 15 \\
        \midrule
        FVD $\downarrow$ & 552.00 & 503.98 & \textbf{489.00} & 510.96 & 510.64 \\
        FID $\downarrow$ & 39.76  & 29.62  & \textbf{26.61}  & 27.55 & 29.67 \\
        \bottomrule
    \end{tabular}
\end{table}

\subsection{Tuning of Static Sampling Baselines}

To further exclude the possibility that our gains come from under-tuned baselines, we perform a systematic grid search over \emph{static} top-$p$, \emph{static} top-$k$, and combined $pk$ sampling on DrivingWorld, while keeping all other settings (including temperature) fixed.
For each baseline family, we report the \emph{best} configuration found in the grid and compare it against ENkG.

\paragraph{Static top-$p$ baselines.}
We sweep over $p \in \{0.5, 0.7, 0.8, 0.9, 1.0\}$.
The quantitative results are summarized in Table~\ref{tab:baseline_topp}.
Even under the best static configuration, the FVD remains above $530$ and FID around $40$, which are clearly worse than ENkG (FVD $=489.00$, FID $=26.61$).

\begin{table}[t]
    \centering
    \caption{Static top-$p$ baselines on DrivingWorld.
    ENkG (default configuration from Table~\ref{tab:enk_sensitivity_thresholds}) is shown for reference.
    Even the best static top-$p$ setup remains substantially worse than ENkG in FVD/FID.}
    \label{tab:baseline_topp}
    \vspace{0.3em}
    \begin{tabular}{l|ccccc}
        \toprule
        Metric & $p=0.5$ & $p=0.7$ & $p=0.8$ & $p=0.9$ & $p=1.0$ \\
        \midrule
        FVD $\downarrow$  & 836.49 & 680.62 & 642.97 & 625.44 & \textbf{530.22} \\
        FID $\downarrow$  & 52.64 & 46.86 & \textbf{40.03} & 47.26 & 43.73 \\
        LPIPS $\downarrow$& 0.40  & \textbf{0.36} & 0.37 & 0.38 & 0.38 \\
        SSIM $\uparrow$   & 0.46  & \textbf{0.50} & 0.46 & 0.43 & 0.34 \\
        PSNR $\uparrow$   & 13.50 & 15.76 & 14.75 & \textbf{16.46} & 14.66 \\
        \bottomrule
    \end{tabular}
\end{table}

\paragraph{Static top-$k$ baselines.}
We further sweep over $k \in \{30, 60, 90, 120, 150, 500\}$.
Table~\ref{tab:baseline_topk} reports the results.
Although FID is minimized around $k=90$ and FVD around $k=150$, even these best-performing static top-$k$ configurations still lag behind ENkG in both FVD and FID.

\begin{table}[t]
    \centering
    \caption{Static top-$k$ baselines on DrivingWorld.
    The best static top-$k$ settings (e.g., $k=90$ for FID, $k=150$ for FVD) remain inferior to ENkG.}
    \label{tab:baseline_topk}
    \vspace{0.3em}
    \begin{tabular}{lcccccc}
        \toprule
        Metric & $k=30$ & $k=60$ & $k=90$ & $k=120$ & $k=150$ & $k=500$ \\
        \midrule
        FVD $\downarrow$  & 696.14 & 661.52 & 615.37 & 569.10 & \textbf{554.74} & 564.10 \\
        FID $\downarrow$  & 61.78 & 41.54 & \textbf{34.50} & 37.86 & 39.02 & 39.10 \\
        LPIPS $\downarrow$& 0.40  & 0.39  & 0.39 & \textbf{0.37} & 0.38 & 0.38 \\
        SSIM $\uparrow$   & 0.44  & 0.45  & 0.44 & \textbf{0.48} & 0.45 & 0.45 \\
        PSNR $\uparrow$   & 14.04 & 14.32 & 14.05 & \textbf{15.73} & 15.34 & 15.08 \\
        \bottomrule
    \end{tabular}
\end{table}

\paragraph{Combined $pk$ baselines.}
Finally, we explore combined $pk$ sampling (first top-$k$, then top-$p$ within the truncated set).
Among the tested configurations, we find the best performance around $(p=0.8, k=1000)$:
\[
    \text{FVD} = 595.93,\quad
    \text{FID} = 43.76,\quad
    \text{LPIPS} = 0.357,\quad
    \text{SSIM} = 0.50,\quad
    \text{PSNR} = 15.93.
\]
While the perceptual metrics (LPIPS/SSIM/PSNR) are comparable to ENkG, the FVD and FID are still notably worse.

\paragraph{Discussion.}
Across all these sweeps, we always compare ENkG against the \emph{best} static configuration of each baseline family (top-$p$, top-$k$, and $pk$).
ENkG consistently achieves substantially better FVD and FID, indicating that its advantage does not come from weak or under-tuned baselines, but from the proposed \emph{entropy-guided dynamic truncation with $k$-guard}, which provides a strictly stronger sampling strategy than any single static choice of $(p)$ or $(k)$.

Interestingly, we also observe that for very large candidate sets (e.g., $p=1.0$ in top-$p$ or $k>100$ in top-$k$), the generated videos can exhibit visibly fragmented or ``shattered'' structures, yet FVD does not necessarily increase and can even improve.
This behavior is consistent with known limitations of FVD in penalizing spatial incoherence, and explains why DrivingWorld's original implementation adopts relatively  defaults (e.g., $k=30$) to preserve vehicle structural consistency rather than aggressively minimizing FVD under heavily fragmented scenes.

\section{Qualitative Results}

\subsection{General-Domain Models}
\begin{figure}[h]
    \centering
    \includegraphics[width=\linewidth]{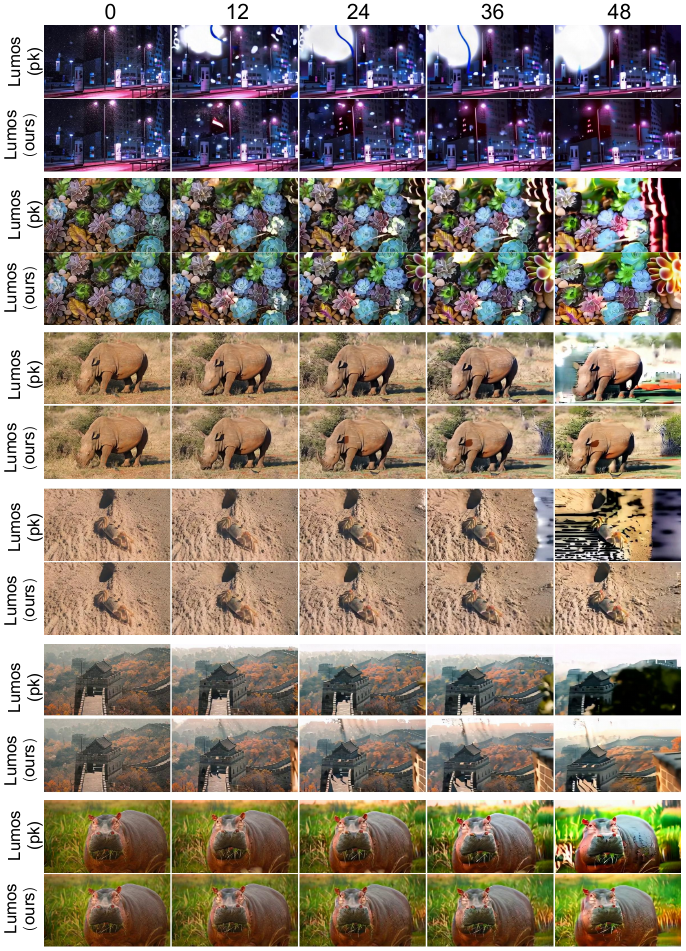}
    \caption{Comparative experiments on the \textit{Lumos-1} model.}
    \label{fig:lumos}
\end{figure}

To validate the effectiveness of ENkG in general domains, we evaluate it on Lumos-1~\citep{yuan2025lumos1autoregressivevideogeneration} and NBP~\citep{ren2025blockpredictionvideogeneration}. Our method mitigates error accumulation during long-horizon video generation and produces more temporally consistent results with reduced color drift and collapse artifacts. Representative qualitative results are shown in Figure~\ref{fig:lumos}.

\subsection{Additional Comparisons}
We further provide more qualitative results on DrivingWorld in Figure~\ref{fig:drivingworld} and VaVim in Figure~\ref{fig:vavim}.

\subsection{Long-Horizon Generation on DrivingWorld}
\begin{figure}[h]
    \centering
    \includegraphics[width=\linewidth]{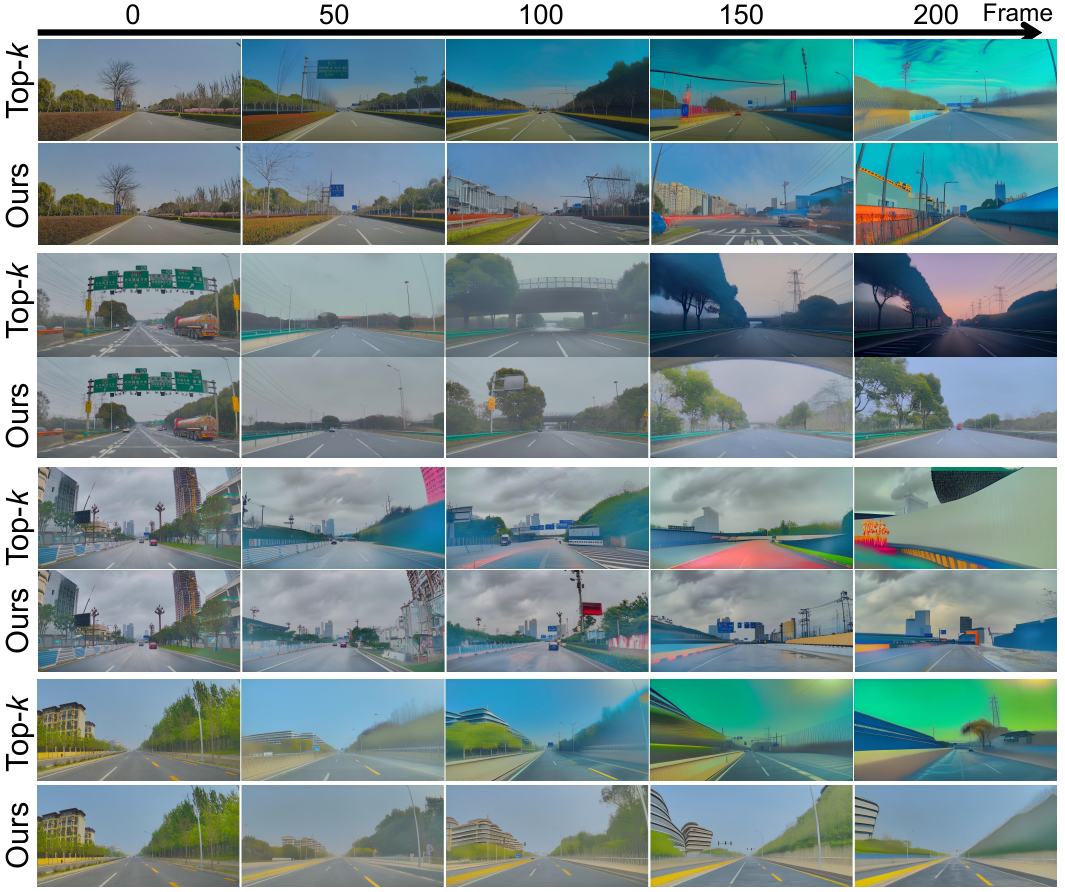}
    \caption{Long-term generation results of Drivingworld.}
    \label{fig:long_horizon}
\end{figure}

To further evaluate ENkG under long-range autoregressive rollout, we conduct
a 200-frame generation experiment on DrivingWorld.
This setting is particularly prone to error accumulation and low-entropy collapse.
As shown in Figure~\ref{fig:long_horizon}, ENkG substantially mitigates visual drift
and maintains scene stability over very long horizons, whereas the baseline model
exhibits background smearing and global color shift.

\section{Analysis of Entropy.}
\subsection{Empirical Examination of Entropy in AR Video Models}
\label{app:entropy_empirical}

\begin{figure}[h] 
    \centering 
    \begin{subfigure}{0.45\textwidth} 
        \centering 
        \includegraphics[width=\linewidth]{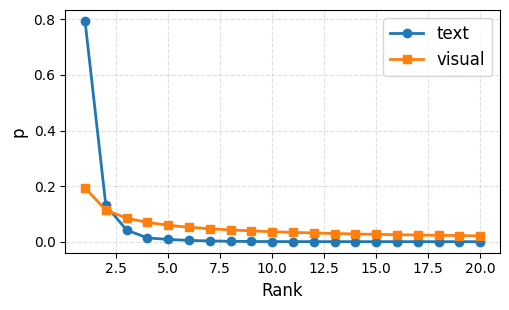} 
        \caption{} 
        \label{fig:entropy_empirical} 
        \vspace{-1em}
    \end{subfigure}
 \hfill
    \begin{subfigure}{0.45\textwidth} 
        \centering 
        \includegraphics[width=\linewidth]{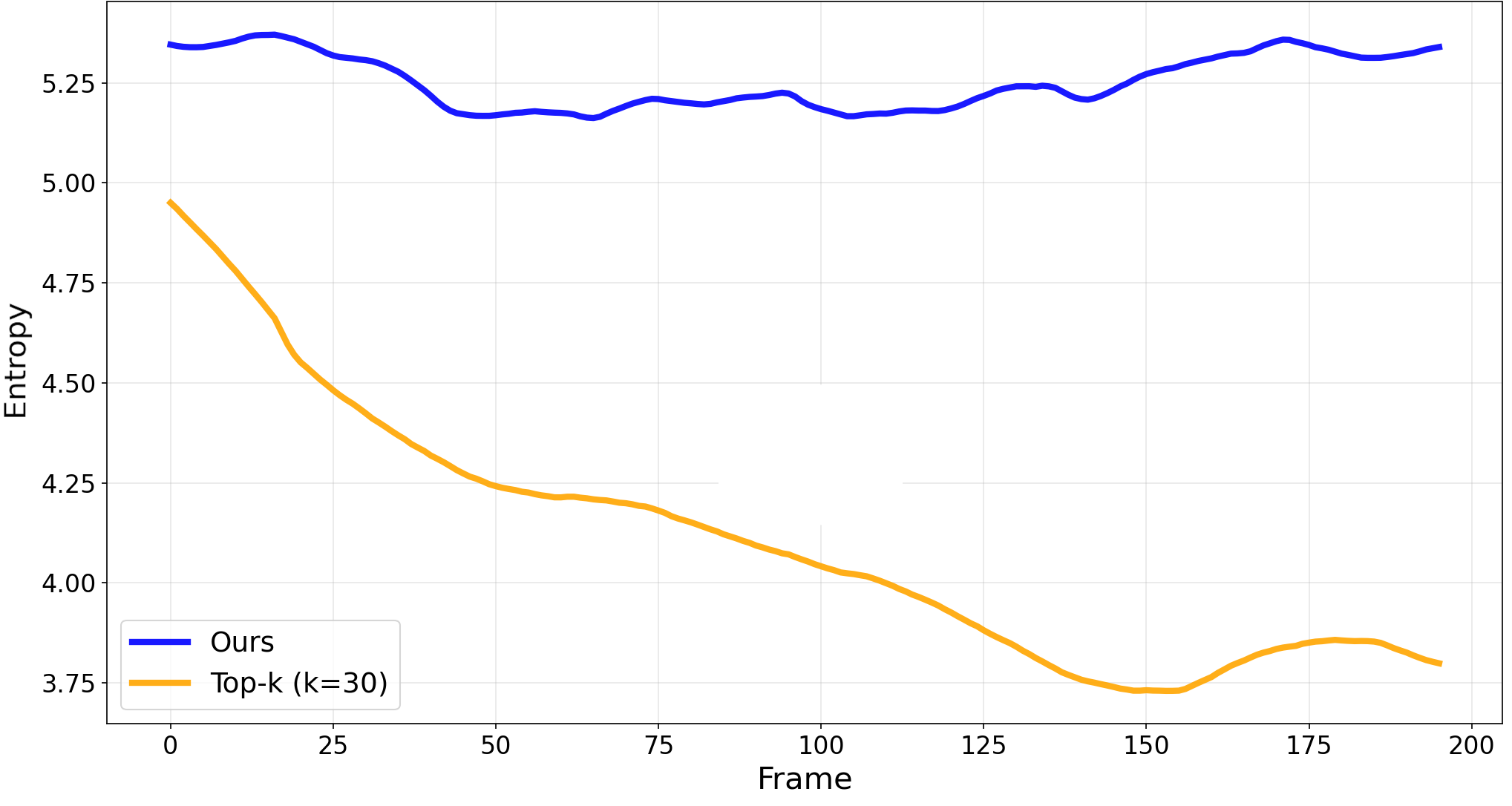} 
        \caption{} 
        \label{fig:entropy_curve} 
        \vspace{-1em}
    \end{subfigure}
    \caption{(a).Comparison of the probabilities of the top-20 tokens between  LLMs (Qwen2.5~\cite{Qwen2.5-VL}) and video AR model (DrivingWorld~\cite{hu2024drivingworld}). (b).Average token entropy of DrivingWorld at each frame as a function of generation timestep on the Nuplan dataset.} 
    \label{fig:two_subfigures} 
\end{figure}

To better understand the entropy dynamics underlying low-entropy collapse,
we compare (a) the probability distributions of the top-20 tokens between a large
language model (Qwen2.5~\cite{Qwen2.5-VL}) and an autoregressive video model 
(DrivingWorld~\cite{hu2024drivingworld}), and (b) the average token entropy across
generation timesteps on the NuPlan dataset. As illustrated in 
Figure~\ref{fig:entropy_empirical}, the output distributions of AR video models are notably flat and diffuse. This observation suggests that video tokens lack direct semantic grounding and exhibit high spatiotemporal redundancy, indicating that semantic information is not localized within individual tokens.

\subsection{Consequences of the Low-Entropy Trap}
\label{app:low_entropy_trap}

In autoregressive (AR) video models, the predictive distribution at step $t$ can be written as
$p_\theta(x_t \mid x_{<t})$ with entropy $H_t = -\sum_x p_\theta(x \mid x_{<t}) \log p_\theta(x \mid x_{<t})$.
We refer to a \emph{low-entropy trap} as the regime where $H_t$ collapses prematurely and
$p_\theta$ becomes pathologically overconfident around a small set of locally consistent tokens, even though the corresponding trajectory is globally suboptimal.

\paragraph{Local overconfidence and trajectory locking.}
Once the model enters such a regime, the effective candidate set under common sampling schemes (top-$p$, top-$k$, or $pk$) often shrinks to one or two tokens with probability mass close to $1$.
From that point on, the model repeatedly feeds its own highly deterministic predictions back into the context, reinforcing the same local pattern step after step.
This behavior is analogous to language models falling into repetitive loops (e.g., ``the the the''), persisting in a wrong chain-of-thought branch, or doubling down on an early but incorrect inference.

\begin{figure}[h] 
    \centering
     \includegraphics[width=0.9\linewidth]{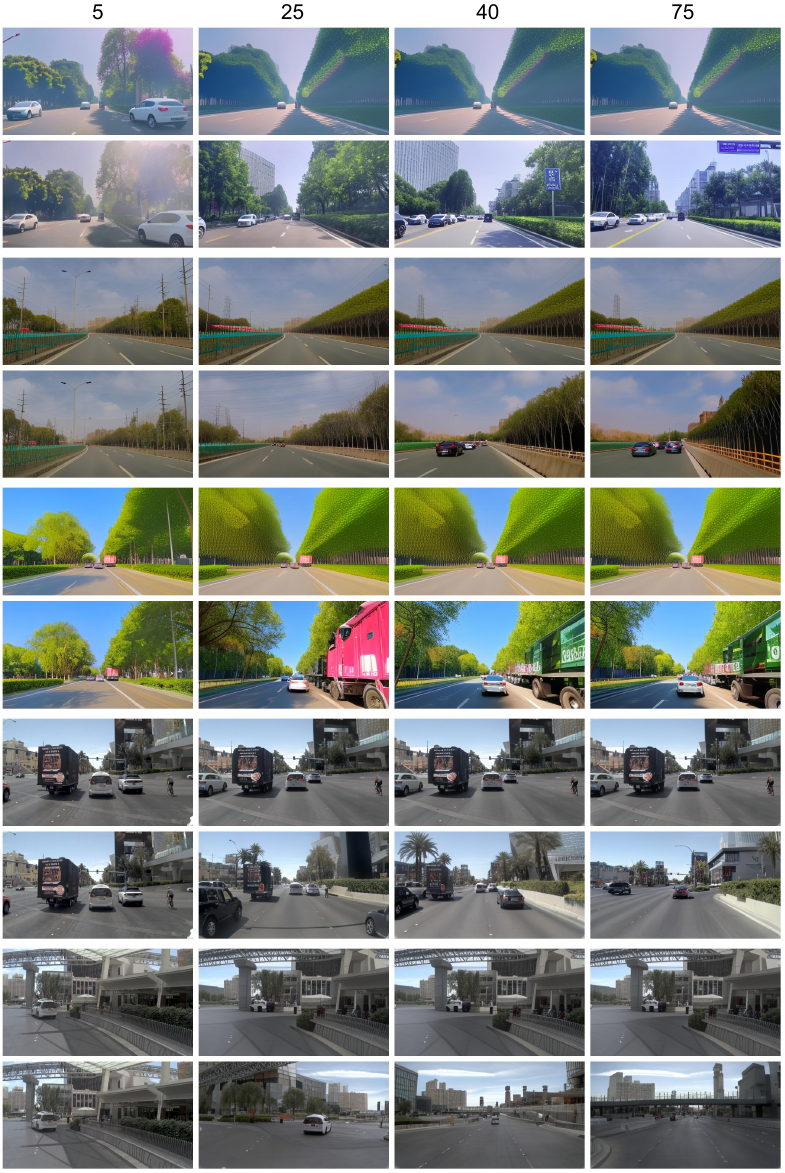}
    \caption{Additional Vavim comparisons between the baseline and ENkG.}
    \label{fig:vavim}
\end{figure}

\paragraph{Visual manifestations in video space.}
In video generation, the low-entropy trap does not merely lead to trivially frozen frames.
More subtly, it manifests as:
(i) \textbf{background smearing}, where large regions of the background collapse into blurry blobs and lose meaningful structural detail;
(ii) \textbf{global color shift}, where the entire frame drifts toward an unnatural color cast, making consecutive frames look consistently tinted or over-/under-exposed; and
(iii) \textbf{texture freezing}, where fine-grained appearance patterns (e.g., grass, water, sky) become unnaturally static and appear glued to the camera or object rather than evolving with the motion.
These effects are illustrated in Figure \ref{fig:drivingworld} and \ref{fig:vavim}, where the model settles into visually coherent yet clearly undesirable trajectories.

\paragraph{Distinction from high-entropy noise.}
It is important to distinguish the low-entropy trap from the more familiar high-entropy failure mode.
High entropy typically corresponds to excessive randomness, producing noisy or chaotic frames that visibly violate short-term consistency.
In contrast, low-entropy collapse yields \emph{overly deterministic} behavior:
short-term consistency can even look improved, while global realism, long-horizon dynamics, and scene plausibility deteriorate.

\paragraph{Effect of $k$-guard.}
Our ENkG sampler directly targets this overconfidence.
When entropy falls below the lower threshold $H_{\text{low}}$, conventional truncation schemes would typically select only the single most probable token.
In contrast, ENkG enforces a minimum guard size $k_g$:
even in very low-entropy regimes, at least $k_g$ candidates are preserved, preventing the effective distribution from collapsing to a point mass.
This mechanism maintains a controlled level of uncertainty and allows the model to explore alternative continuations that can recover from early mistakes.
Empirically, we observe that enabling $k$-guard reduces the frequency and severity of texture locking, background freezing, and unnatural motion, leading to more coherent long-horizon trajectories and higher overall generative quality.

\begin{figure}[h]
    \centering
    \includegraphics[width=0.9\linewidth]{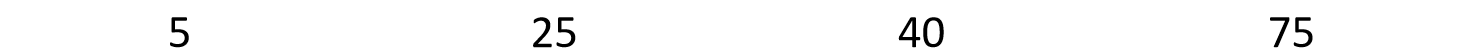}
    \includegraphics[width=0.9\linewidth]{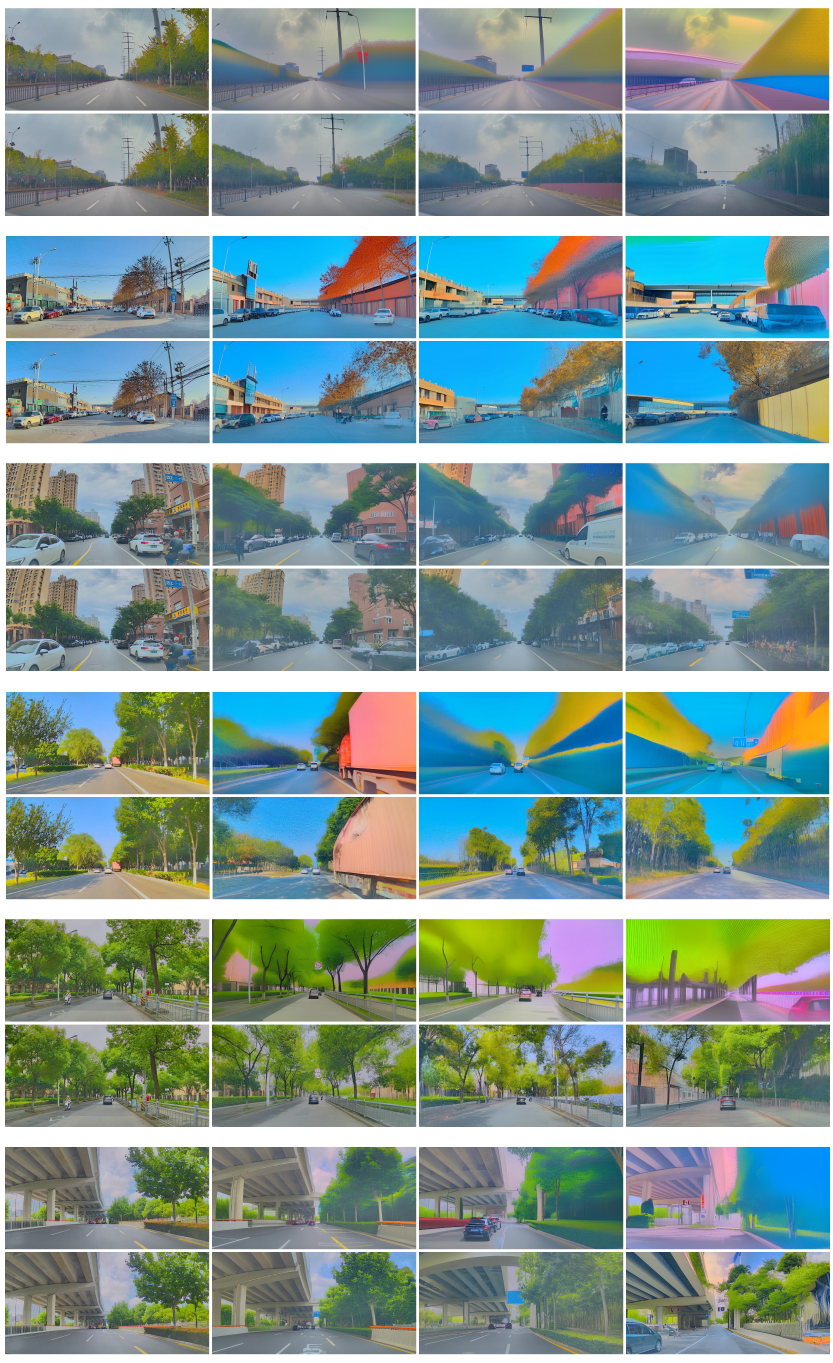}
    \caption{Additional DrivingWorld comparisons between the baseline and ENkG. (Part1)}

    \label{fig:drivingworld}
\end{figure}

\begin{figure}[h] \ContinuedFloat
    \centering
     \includegraphics[width=0.9\linewidth]{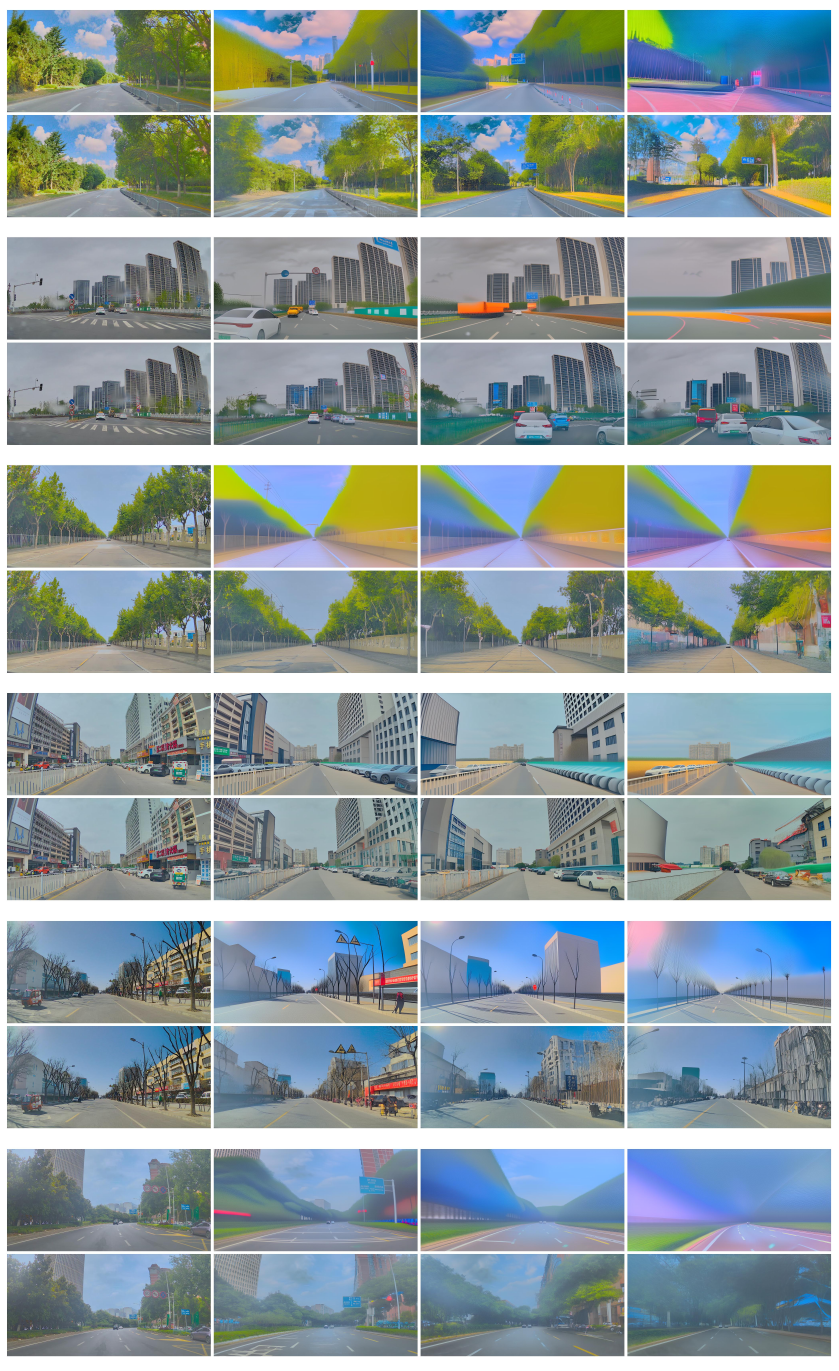}
    \caption{Part2}
\end{figure}

\end{document}